\pdfoutput=1

\documentclass[11pt]{article}

\usepackage{verbatim}
\usepackage{fancyvrb}
\usepackage{multicol}
\usepackage{listings}
\usepackage[final]{acl}

\usepackage{times}
\usepackage{latexsym}

\usepackage[T1]{fontenc}

\usepackage[utf8]{inputenc}

\usepackage{microtype}

\usepackage{inconsolata}

\usepackage{graphicx}

\usepackage{hyperref}       
\usepackage{url}            
\usepackage{booktabs}       
\usepackage{amsfonts}       
\usepackage{nicefrac}       
\usepackage{microtype}      
\usepackage{xcolor}         

\usepackage{hyperref}
\usepackage{url}
\usepackage{tikz}
\usepackage{multirow}
\usepackage{xspace}
\usepackage{booktabs}
\usepackage{ulem}
\usepackage{amsmath}
\usepackage{graphicx} 

\usepackage{wrapfig}

\usepackage{caption}
\usepackage{adjustbox}

\usepackage[flushleft]{threeparttable}
\usepackage{enumitem}
\usepackage{verbatim}
\usepackage{listings}
\usepackage{xcolor}
\usepackage{amssymb}
\usepackage{scalerel} 
\usepackage{float}
\usepackage{subcaption}
\usepackage[inkscapeformat=png]{svg}
\usepackage{bbding}
\usepackage{comment}
\usepackage{tabularray}
\usepackage{booktabs,tabularx}
\usepackage{stfloats}

\newcommand{\approach}{\texttt{SaySelf}\xspace}

\definecolor{ao(english)}{rgb}{0.0, 0.5, 0.0}

\NewDocumentCommand{\xingyao}
{ mO{} }{\textcolor{orange}{\textsuperscript{\textit{Xingyao}}\textsf{\textbf{\small[#1]}}}}

\NewDocumentCommand{\yy}
{ mO{} }{\textcolor{pink}{\textsuperscript{\textit{coolYY}}\textsf{\textbf{\small[#1]}}}}

\NewDocumentCommand{\shizhe}
{ mO{} }{\textcolor{blue}{\textsuperscript{\textit{shizhe}}\textsf{\textbf{\small[#1]}}}}

\NewDocumentCommand{\tianyang}
{ mO{} }{\textcolor{green}{\textsuperscript{\textit{tianyang}}\textsf{\textbf{\small[#1]}}}}

\usepackage{microtype}
\usepackage{hyperref}

\title{\includegraphics[width=0.04\textwidth]{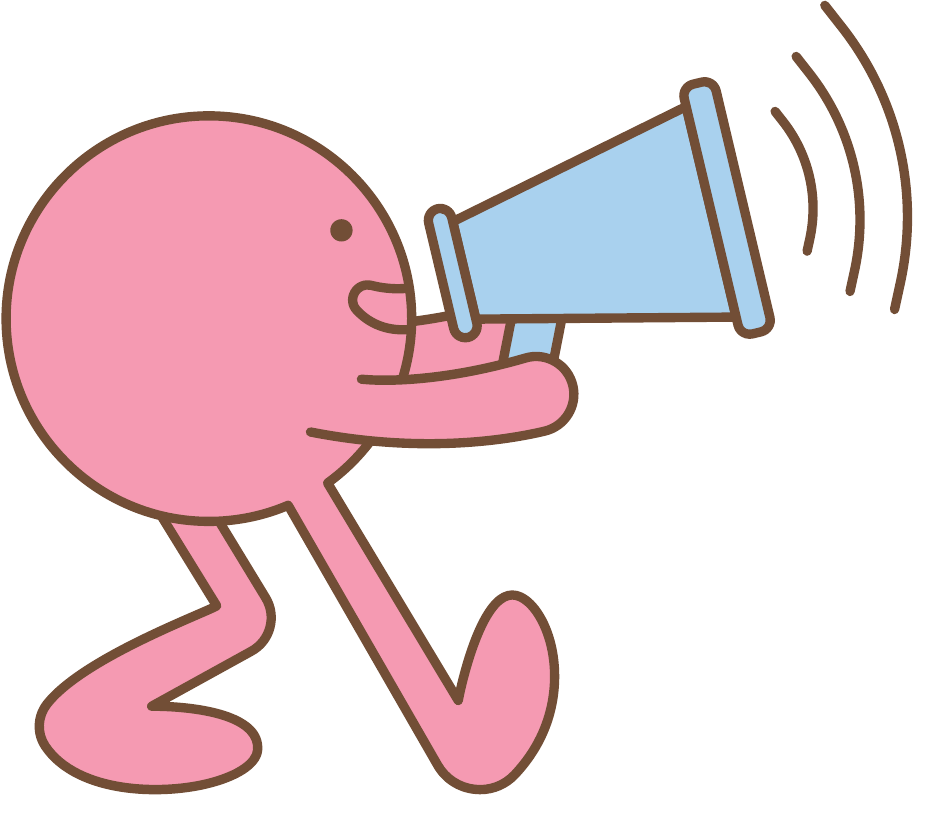}\approach: Teaching LLMs to Express Confidence \\ with Self-Reflective  Rationales}

\author{%
  Tianyang Xu$^{1}$\thanks{Equal contribution.}, Shujin Wu$^{3*}$, Shizhe Diao$^{4}$, Xiaoze Liu$^{1}$\\
   \textbf{Xingyao Wang}${^{2}}$, \textbf{Yangyi Chen}$^{2}$\thanks{Corresponding author.}, \textbf{Jing Gao}$^{1\dag}$ \\
  $^{1}$ Purdue University $^{2}$ University of Illinois Urbana-Champaign \\
  $^{3}$ University of Southern California 
  $^{4}$ NVIDIA \\
  \texttt{\{xu1868, xiaoze, jinggao\}@purdue.edu}, \texttt{shujinwu@usc.edu} \\
  \texttt{sdiao@nvidia.com},
  \texttt{\{xingyao6,yangyic3\}@illinois.edu}
}

\begin{document}

\maketitle

\begin{abstract}
\looseness=-1
Large language models (LLMs) often generate inaccurate or fabricated information and generally fail to indicate their confidence, which limits their broader applications. Previous work has elicited confidence from LLMs by direct or self-consistency prompting, or constructing specific datasets for supervised finetuning. The prompting-based approaches have inferior performance, and the training-based approaches are limited to binary or inaccurate group-level confidence estimates. In this work, we present \approach, a novel training framework that teaches LLMs to express more fine-grained confidence estimates. In addition, beyond the confidence scores, \approach initiates the process of directing LLMs to produce self-reflective rationales that clearly identify gaps in their parametric knowledge and explain their uncertainty. This is achieved by using an LLM to automatically summarize the uncertainties in specific knowledge via natural language. The summarization is based on the analysis of the inconsistency in multiple sampled reasoning chains, and the resulting data is utilized for supervised fine-tuning. Moreover, we utilize reinforcement learning with a meticulously crafted reward function to calibrate the confidence estimates, motivating LLMs to deliver accurate, high-confidence predictions and to penalize overconfidence in erroneous outputs. Experimental results demonstrate the effectiveness of \approach in reducing the confidence calibration error and maintaining the task performance. 
The generated self-reflective rationales are also reasonable and can further contribute to the calibration. 
The code is made public at \url{https://github.com/xu1868/SaySelf}.


%

%
%

%

\end{abstract}

\section{Introduction}\label{sec:intro}

While large language models (LLMs) exhibit remarkable proficiency in reasoning and generating effective responses~\cite{openai2023gpt4, touvron2023llama, jiang2023mistral, wang2024executable},
they often produce fabricated information (\textit{a.k.a,} hallucination) and typically hesitate to indicate their uncertainty when faced with unfamiliar questions~\cite{ye2023cognitive, liu2023prudent}. 
Determining how to accurately obtain reliable confidence estimates from LLMs is essential~\cite{xiong2023can, zhou2023navigating}, particularly when the responses are not limited to single tokens\footnote{The confidence of a single-token answer can be derived from its token probability, which typically exhibits high calibration~\cite{openai2023gpt4, Chen2022ACL}.}.

\begin{figure*}[t!]
\centering
\includegraphics[width=\textwidth]{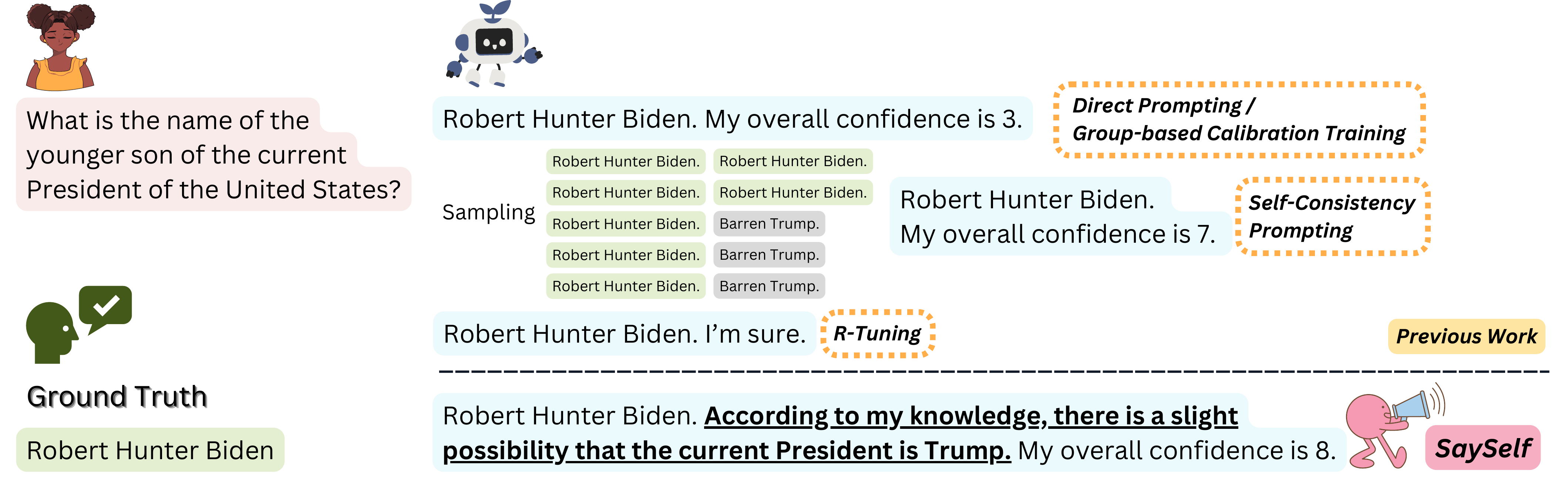}
 \caption{The comparison between \approach and previous work. 
 \approach can produce the \underline{\textbf{self-reflective rationale}} that explains why the model is uncertain and the fine-grained and accurate confidence estimates.
This simple example is constructed for illustration purposes, and the reasoning chain is omitted for brevity.}
 \label{fig:introfigure}
 \end{figure*}

Previous work on eliciting confidence from LLMs includes prompting-based and training-based approaches. Prompting-based methods, such as direct prompting and self-consistency prompting in Figure \ref{fig:introfigure}, employ specific prompts to generate confidence scores or use answer consistency as a confidence indicator, even though these can have poor calibration performance or significantly increase inference latency~\cite{tian2023just, xiong2023can, diao2023active}. 
Training-based approaches, such as group-based calibration training and R-Tuning in Figure \ref{fig:introfigure}, develop specialized datasets for fine-tuning that encourage LLMs to express confidence. However, these methods often provide suboptimal or binary confidence estimates, failing to accurately reflect the models' confidence levels~\cite{lin2022teaching, zhang2023r, yang2023alignment}. In conclusion, previous work tends to suffer from the following problems: (1) Poor calibration performance; (2) Coarse-grained confidence levels; (3) Long inference latencies.

%

\looseness=-1
In this work, we present \approach, a training framework that teaches LLMs to generate more accurate and fine-grained confidence estimates. It successfully tackles the aforementioned problems in previous work. 
More importantly, \approach also goes beyond the confidence elicitation in previous work, and further enables LLMs to generate self-reflective rationales that indicate their knowledge gap and explain their confidence estimates (Figure~\ref{fig:introfigure}). 
%
We accomplish this by automatically generating a model-specific dataset for supervised fine-tuning using an off-the-shelf LLM (\textit{e.g.,} GPT-4~\cite{openai2023gpt4}). 
Specifically, for each question, we sample multiple reasoning chains from LLMs. We then perform clustering of the reasoning chains based on the semantic similarity and retain one instance per cluster. 
GPT-4 is then tasked with analyzing these instances from various clusters, summarizing uncertainties in natural language from a first-person perspective, which is subsequently used for fine-tuning.

%

%

For accurate and fine-grained confidence estimates, we employ reinforcement learning to calibrate LLMs' confidence estimate in each response. 
We design a reward function that incentivizes LLMs to produce accurate, high-confidence predictions and imposes penalties for overconfidence in incorrect responses.
In addition, the self-reflective rationales and confidence estimates are generated without multiple sampling, significantly reducing the inference time.


%


\looseness=-1
We evaluate \approach on multiple knowledge-extensive question-answering tasks.
%
We show that \approach significantly reduces the confidence calibration error and maintains the task performance. 
The generated self-reflective rationales effectively capture the internal uncertainty and can further improve the calibration performance.
In addition, we find that \approach enables LLMs to produce low confidence in unanswerable questions. 
%

%

%

Our research has the potential to exert influence on both related academic research and real-world applications, including but not limited to the following cases:
(1) A clear confidence expression with explanations can promote trustworthiness in AI, from the perspective of LLMs' alignment. 
(2) The self-reflective rationales can guide LLMs to perform subsequent steps, like invoking external tools or asking clarification questions, for better interaction and performance. 
(3) We also anticipate promising developments in training protocols once LLMs are trained with \approach, including proactive learning algorithms that enhance LLMs' interactions with humans for continued learning.


%
%

%

%

\section{Related Work}
\paragraph{LLMs' Confidence Elicitation}
Eliciting accurate confidence estimates for LLM-generated answers that contain multiple tokens is challenging~\cite{borji2023categorical, zhou2024relying}. 
Previous work can be categorized into prompting-based and training-based approaches. 
Prompting-based approaches use a specific prompt to guide LLMs to generate confidence scores for their predictions~\cite{tian2023just, kadavath2022language}, or prompt LLMs to generate the answers multiple times and use the consistency levels as indicators of their confidence~\cite{xiong2023can, lyu2024calibrating, diao2023active, yang2023alignment}.
These approaches can cause inferior performance or lead to extensive inference-time latency.
Training-based approaches construct a specialized dataset for supervised fine-tuning, encouraging LLMs to express their uncertainty. 
\citet{lin2022teaching} first group examples based on their types of question as the label, then obtains the confidence score for each example using the empirical accuracy for the whole group.
This approach can lead to suboptimal confidence estimates since not all examples in the same group are equal.
%
%
R-Tuning~\cite{zhang2023r} reconstructs the SFT data to add ``I am sure/unsure'' at the end of the correct/incorrect responses, which can only generate binary uncertainty estimates.
As mentioned in Section \ref{sec:intro}, \approach addresses the limitations of the previous methods and guides LLMs to generate more accurate and fine-grained confidence estimates.

\paragraph{LLMs' Hallucination \& Uncertainty Expression}
LLMs' hallucination refers to instances where these models generate information that is not supported by their training data or the input provided~\cite{Zhang2023SirensSI, liang2024learning, agrawal2023language}. 
Numerous research is dedicated to exploring the causes of hallucination \cite{Dziri2022OnTO, McKenna2023SourcesOH, DBLP:journals/corr/abs-2402-03757} and developing methods to detect or mitigate hallucination~\cite{varshney2023stitch, Rawte2023ASO, Andriopoulos2023AugmentingLW}.
Besides the hallucination, the reluctance of LLMs to express uncertainty when they are unable to solve tasks can further erode trust in these systems~\cite{ji2023ai, zhou2024relying}.
Existing research identifies the tendency in LLMs to fabricate information when addressing unknown questions~\cite{liu2023prudent, hu2023won, amayuelas2023knowledge}.
This inability can be traced back to the supervised instruction finetuning (SFT) stage, which trains LLMs on human-written or GPT-synthesized (instruction, response) pairs~\cite{wang2022self, peng2023instruction}. 
This paradigm neglects the discrepancy between pretraining and SFT data, potentially inducing hallucinations by instructing LLMs to appear helpful, even when they are unable to solve the problem, and discouraging them from expressing uncertainty or declining responses~\cite{zhang2023r}.
In this work, we propose \approach to train LLMs to express accurate confidence estimates and self-reflective rationales as an efficient method against hallucinations, as they can guide end users to verify information in responses and help them regain confidence in LLMs.
%

\paragraph{LLMs' Explainability}
Our work is also related to explainability for LLMs regarding the self-reflective rationales generation~\cite{zhao2024explainability, singh2024rethinking}. 
Previous work on natural language explanations for LLMs 
provides motivation to explain the models' decision-making process for a prediction~\cite{costa2018automatic, cambria2023survey}. 
The typical approaches to producing natural language explanations involve training LLMs with the ground-truth labels and the human-annotated explanations that can serve as effective augmented supervision that guide LLMs to reason in a right way~\cite{rajani2019explain, luo2021local, yordanov2021few}.
Another line of research adopts chain-of-thought (CoT) reasoning as natural language explanations~\cite{wei2022chain, lyu2023faithful, xu2024chain, chen2023measuring}. 
Compared to previous methods, \approach significantly departs from existing methods by generating rationales that not only justify the predictions but also elucidate the confidence estimates.
Most importantly, \approach adopts LLMs' internal reasoning process to generate self-reflective rationales, instead of human-annotated explanations, which may not be faithful to specific LLMs.
Unlike CoT, which primarily clarifies the rationale behind predictions, \approach also explicates the sources of uncertainty.

\section{\approach}

 \begin{figure*}[t!]
\centering
\includegraphics[width=\textwidth]{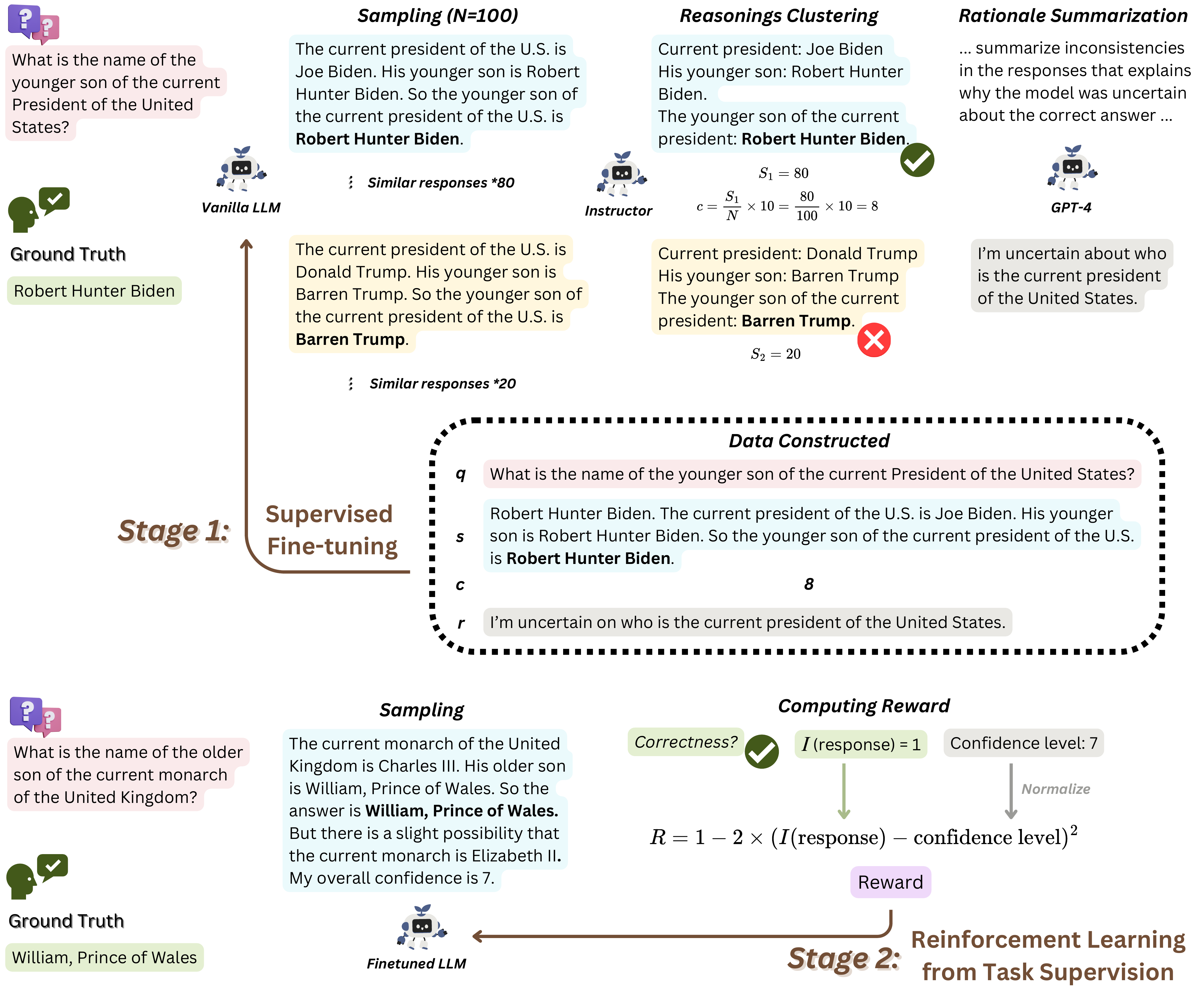}
 \caption{The overview of \approach, consisting of the supervised fine-tuning and reinforcement learning from task supervision stages. The former stage trains LLMs to generate self-reflective rationales and confidence estimates based on multiple sampling, and the latter stage employs reinforcement learning to further calibrate the confidence estimates based on task supervision. $q$, $s$, $c$, and $r$ denote question, response, confidence estimate, and self-reflective rationale respectively.
 }
 \label{fig:approach}
 \end{figure*}

We present \approach, a training framework to teach LLMs to express fine-grained confidence with self-reflective rationales (see Figure~\ref{fig:approach}). 
\approach consists of 2 essential stages:
(1) \textbf{Supervised Fine-Tuning}: 
We establish a model-specific dataset containing self-reflective rationales and confidence estimates. This dataset is built from multiple sampled responses from LLMs.
(2) \textbf{Reinforcement Learning from Task Supervision}:
We use reinforcement learning with a carefully designed reward function to further calibrate the confidence estimates for each instance. 
For both 2 stages, we adopt the training samples in HotpotQA~\cite{yang2018hotpotqa}, which typically require multi-step reasoning on knowledge facts to derive the answer.
After the two-stage training, the trained models can directly answer questions with confidence estimates and self-reflective rationales without additional computational overhead.



\subsection{Supervised Fine-Tuning}
%
In this stage, our goal is to construct a supervised dataset $D$, where each sample contains a question $q$, an answer with the reasoning chain $s$, the self-reflective rationale $r$, and the confidence estimate $c$. 
Basically, $r$ summarizes specific knowledge that the LLM is uncertain about, and is generated by analyzing the inconsistency in multiple selective responses sampled from the vanilla LLM $M$.
$c$ is an integer from 1 to 10, and is derived based on the consistency of $s$.

We adopt 90K questions in HotpotQA. 
For each question, we prompt $M$ to generate the reasoning chain and the answer $N$ times.
We perform clustering on the $N$ responses to obtain $K$ representative clusters based on the semantic similarity among responses since there is significant redundancy.
Specifically, we adopt the Instructor~\cite{su2022one}, an instruction-finetuned text embedding model that produces text embeddings customized to the specific task and domain.
%
Our clustering process involves examining each response, 
identifying those within the similarity threshold $T$, and grouping them accordingly until all responses have been processed.
The cluster size $S$ is defined as the number of responses in the cluster.
We randomly pick one selected response per cluster for the following steps, as empirical evidence suggests significant similarity among responses within the same cluster (see Appendix~\ref{sec:appb} for details).


\looseness=-1
To derive the confidence estimate $c$, we first check the correctness of the selected response from each group using the golden answer annotated in HotpotQA. 
Samples with no correct responses are removed to avoid training LLMs on incorrect examples.
The correct response is taken as the golden $s$ for this sample, and $c$ is computed as: $c=round(\frac{S_c}{N} * 10)$, where $S_c$ is the size of $s$'s cluster, and $round(x)$ returns the nearest integer of $x$.

To derive the self-reflective rationale $r$, we instruct GPT-4 to carefully analyze and compare all selected responses, focusing on the inconsistency in the provided knowledge facts.
Then GPT-4 is required to summarize ``why $M$ is uncertain'' in natural language from a first-person perspective.
The summary is thus taken as the self-reflective rationale $r$.
The prompt is provided in Appendix~\ref{app:prompt}.

We train the vanilla $M$ on $D$ via supervised fine-tuning. The objective function is:
\begin{align}
&\max_\Theta \sum_{(q,s,r,c') \in D} [\  \log P(s | q; \Theta) \nonumber\\ 
&+ \log P(r | s, q; \Theta) + \log P(c' | s, r, q; \Theta) ]
\end{align}
where \( \Theta \) represents the parameters of $M$, $c'$ is the natural language expression of the confidence estimate $c$ (\textit{a.k.a.,} ``My confidence is $c$''). The objective function is meant to maximize the sum of these log probabilities over all the tuples $(q,s,r,c')$ in the dataset $D$.
%

%
%
%
%
%

\subsection{Reinforcement Learning from Task Supervision}
%
\looseness=-1
Due to the nature of supervised fine-tuning, the model tends to produce homogeneous confidence levels, such as relatively lower confidence levels for correct responses and higher levels for incorrect responses. To address this issue, we use reinforcement learning to further calibrate LLMs' fine-grained confidence estimates and guide the model to produce more accurate and differentiated values. 
%
%
During the sampling phase, LLMs are prompted to produce responses, self-reflective rationales, and confidence levels.
To optimize the model, we compare the generated response with the ground truth.
Subsequently, we formulate a reward function considering answer accuracy and model confidence. To encourage the model towards more differentiated values, the reward function has a quadratic output: 
\begin{equation}\label{eq:reward}
    R = 1 - 2 * (\mathbb{I}(\textrm{response}) - \textrm{confidence level}) ^ 2
\end{equation}
where $\mathbb{I}()$ is the indicator function, which returns 1 if the generated response is correct, else 0. 
The confidence level is normalized between 0 and 1.
This reward function reinforces LLMs for high confidence in accurate samples while penalizing them for being overconfident in incorrect ones.

We utilize the Proximal Policy Optimization (PPO) algorithm~\cite{schulman2017proximal} to train LLMs based on this defined reward function. 
The optimization objective is expressed as:
\begin{align}
\max_{\Theta} \ \  \mathbb{E}_{t} [ &\min(r_t(\Theta) \hat{A}_t, \text{clip}(r_t(\Theta),\nonumber 
\\ &1 - \epsilon, 1 + \epsilon) \hat{A}_t)]
\end{align}
where \( r_t(\Theta) \) calculates the probability ratio of the newly proposed policy to the old policy.
The advantage estimate \( \hat{A}_t \), crucial for directing updates, is calculated from the difference between the anticipated future rewards under the current policy and the baseline or value function.
This advantage estimate is directly influenced by the reward \( R \), which in turn ties the optimization process closely with both response accuracy and confidence level.

\subsection{Implementation Details}
For the supervised dataset collection, the sampling time $N$ is set to 100 and the temperature is 1.2. 
The similarity threshold $T$ is set to 0.9. 
For supervised fine-tuning, the learning rate is set to 7e-5 and the batch size is set to 8.
For the reinforcement learning stage, the learning rate is set to 1e-5 and the batch size is set to 8. 
To check the correctness of the responses, we utilize a verification method where annotated answers must be present within the responses. This heuristic demonstrates high precision in knowledge-based QA tasks.
%



%
%

\section{Experiments}

%
%

\subsection{Evaluation Setting}\label{sec:evaluation_setting}
\looseness=-1
\paragraph{Evaluation Datasets}
We follow \citet{zhang2023r} to evaluate LLMs on knowledge-extensive QA tasks.
We include the following datasets:
\textbf{HotpotQA}~\cite{yang2018hotpotqa}, a dataset of multi-hop reasoning question-answer pairs;
\textbf{TruthfulQA}~\cite{lin2021truthfulqa}, a dataset that tests whether models generate truthful answers to questions specifically designed to induce false answers;
\textbf{StrategyQA}~\cite{geva2021did}, a dataset of true/false questions requiring multi-hop reasoning;
\textbf{FEVER}~\cite{thorne2018fever}, a dataset used to assess the ability of models to verify the factuality of statements against Wikipedia documents;
\textbf{HaluEval}~\cite{li2023halueval}, a dataset that evaluates the hallucination of models;
\textbf{ParaRel}~\cite{elazar2021measuring}, a dataset that measures the model's performance in understanding paraphrased relational facts.

\paragraph{Evaluation Environments}
The experiments are run on a server with 4 Nvidia A6000 GPUs and 256GB RAM. The models are implemented with the Huggingface Transformers (\url{https://huggingface.co/}) library. The reported data are all average values of three runs. Both stages take approximately 1 hour to train during the two-stage training process.
%
\paragraph{Evaluation Metrics}
We measure various approaches from 3 aspects.
(1) \textbf{Confidence Calibration Performance}: 
We adopt 2 calibration metrics. First, we use the ECE score to measure the confidence calibration error~\cite{guo2017calibration, Chen2022ACL}. 
Basically, ECE evaluates the correlation between the confidence scores assigned by LLMs and their corresponding correctness. For responses from LLMs $A$, it can be calculated as
\begin{equation}
    \mathrm{ECE}=\frac{1}{|A|} \sum_{a\in A}|\mathbb{I}(a)-\mathrm{conf}(a)|, 
\end{equation}
where $\mathbb{I}()$ is the indicator function defined in Equation \ref{eq:reward}, and $\mathrm{conf}()$ returns the confidence level of LLMs.
Second, we adopt the AUROC score following \cite{hendrycks2016baseline}. 
It measures the ability of LLMs to distinguish between correct and incorrect responses across different threshold settings. It can be calculated as
\begin{equation}
    \mathrm{AUROC} = \int_0^1 \mathrm{TPR}(\mathrm{FPR^{-1}}(x)) \mathrm{d}x,
\end{equation}
where $x$ is the threshold confidence level, $\mathrm{TPR}$ is the true positive rate under this threshold confidence level, and $\mathrm{FPR}$ is the false positive rate under the threshold.
(2) \textbf{Task Performance}: We measure the typical accuracy on the test split of the datasets. 
(3) \textbf{Faithfulness of the Generated Self-Reflective Rationales}: 
We make the first effort to measure the faithfulness of the provided self-reflective rationales.
We suggest employing the same intuition utilized in \approach.
For each question, we sample multiple responses (answers with reasoning chains) from the LLM, and perform clustering to retain several representative responses.
Subsequently, we utilize a proficient LLM (GPT-4) to examine whether the provided self-reflective rationales can faithfully express the uncertainty demonstrated in the sampled responses, and give a score from 1 to 10. 
The final faithfulness score is the average over all samples.
%

%


%
%

%

%

\begin{table*}[]
\centering
\renewcommand\arraystretch{0.8}
\setlength{\tabcolsep}{9pt} 

\resizebox{0.99\textwidth}{!}{
\begin{tabular}{l|cccccc}
\toprule
Method | Dataset  & HotpotQA       & TruthfulQA      & StrategyQA      & FEVER           & HaluEval        & ParaRel         \\ \midrule
DP                & 0.6667          & 0.3437          & 0.5357          & 0.4529          & 0.6746          & 0.5129          \\
SC                & 0.3830          & 0.5204          & 0.3957          & 0.4537          & 0.4242          & 0.5458          \\
PC               & 0.5515      & 	0.4963	              & 0.4379	         & 0.4659	    & 0.3080	              & 0.5071          \\

R-Tuning          &  0.4141	        &  0.4111	                  &         0.4477         &    	0.4007	             &              0.2777	 &        0.6797              \\
AS          &  0.3833	          & 0.4308	& 0.4125	          & 0.3973	     & 0.4344	           & 0.3926            \\

GCE               &   0.3597	      &           0.3639	          &  0.4474                  & 	0.4473	                 &      0.5819	       &        0.4634              \\ \midrule
\approach           & \textbf{0.3558}* & \textbf{0.3368}* & \textbf{0.3907}* & \textbf{0.3704}* & \textbf{0.2661}* & \textbf{0.3272}* \\
w/o RL            & 0.3704          & 0.3887          & 0.3951          & 0.3903          & 0.2804          & 0.3628          \\
w/o R \& CE & 0.5063          & 0.4286          & 0.4195          & 0.4313          & 0.4143          & 0.3972         \\ 
w/o R & 0.3750	& 0.3609 &	0.3938 &	0.3854 &	0.4294 &	0.4730 \\
w/ Naive RF & 0.6129	& 0.4356 &	0.4062	 & 0.4238 &	0.2812	 & 0.3316       \\ 

\bottomrule
\end{tabular}

}
\caption{The ECE evaluation results of baselines, \approach, and various ablations. 
Lower is better. 
HotpotQA is the only in-distribution dataset.
$p$-Values are the $p$-values comparing \approach over other methods.
In this table, \textbf{DP} denotes direction prompting, \textbf{SC} denotes self-consistency, \textbf{PC} denotes prompting for correctness, \textbf{AS} denotes aligning with self-consistency-based confidence, \textbf{GCE} denotes grouping-based confidence estimates for calibration training; \textbf{w/o RL} denotes \approach without reinforcement learning, \textbf{w/o R \& CE} denotes \approach without self-reflective rationales and confidence estimates, \textbf{w/o R} denotes \approach without self-reflective rationales, \textbf{w/ Naive RF} denotes using another naive reward function. The numbers with asterisk marks (*) mean significant advantage with the statistical significance threshold of $p$-value 0.05 in the paired t-test comparing with baselines.
}

\label{tab:ece}
\end{table*}

\begin{table*}[t!]
\centering
\renewcommand\arraystretch{0.8}
\setlength{\tabcolsep}{9pt} 

\resizebox{0.99\textwidth}{!}{
\begin{tabular}{l|cccccc}
\toprule
Method | Dataset  & HotpotQA       & TruthfulQA      & StrategyQA                    & FEVER           & HaluEval        & ParaRel    \\ \midrule
DP                & 0.1562          & 0.5125          & 0.3904                        & \textbf{0.5713} & 0.4650          & 0.3971          \\
SC                & 0.3288          & \textbf{0.5777} & {\color[HTML]{000000} 0.3697} & 0.5578          & 0.8498 & \textbf{0.6631} \\
PC                & 0.3281 &	0.5546 &	0.4450 &	0.4968 &	0.7012 &	0.4841 \\
R-Tuning          &  0.3664	        &  0.5216                    &        	0.5318	                      &    0.5530	        &         0.8125       &       	0.1430                   \\
AS          & 0.3379	        & 0.4861	& 0.3670	          & 0.4539	          & 0.7547	                     & 0.5684                  \\
GCE               &   0.3635	   &        0.4425	                   &     0.5504                        &         	0.5506	         &  0.8074	           &         0.6168              \\ \midrule
\approach           & 0.3585          & 0.5353          & \textbf{0.5956}               & 0.5393          & 0.8425          & 0.6319          \\
w/o RL            & 0.3708 & 0.4667          & 0.5340                         & 0.5523          & \textbf{0.8527}          & 0.6198          \\
w/o R \& CE & 0.3411          & 0.4623          & 0.3811                        & 0.4004          & 0.7198          & 0.5373          \\
w/o R   &0.3650 &	0.4964 &	0.4224 &	0.4848	 & 0.7652 & 	0.5707   \\
w/ Naive RF & \textbf{0.3715} &	0.5721	& 0.5811 &	0.5443 &	0.8450	& 0.6577     \\ 
\bottomrule
\end{tabular}
}
\caption{The accuracy evaluation results of baselines, \approach, and various ablations.}

\label{tab:acc}
\end{table*}

\begin{table*}[t!]
\centering
\renewcommand\arraystretch{0.8}
\setlength{\tabcolsep}{9pt} 

\resizebox{0.99\textwidth}{!}{
\begin{tabular}{l|cccccc}
\toprule
Method | Dataset & \multicolumn{1}{l}{HotpotQA} & \multicolumn{1}{l}{TruthfulQA} & \multicolumn{1}{l}{StrategyQA} & \multicolumn{1}{l}{FEVER} & \multicolumn{1}{l}{HaluEval} & \multicolumn{1}{l}{ParaRel}\\ \midrule
DP / SC          & 6.5                                   & 7.8                                     & 5.9                                     & 6.2                                & 7.5                                   & 7.0                                  \\
PC         & 6.4 &	7.3 &	5.2	 &6.0 &	7.1 &	7.2  \\
R-Tuning          & 6.7	                                   & 7.4	                             & 6.0	                                      & 6.2                         & 	6.7	                                  & 6.1      \\ 
AS          & 5.1	                                   & 6.6                             & \textbf{8.1}                                      & 7.6                         & 8.0	                                  & 6.6      \\  
GCE          & 5.7	& 6.1	& 4.2	& 5.6 &	5.9	& 5.2  \\
\approach          & \textbf{8.3}                          & \textbf{8.6}                            & {\color[HTML]{000000} 5.5}     & \textbf{7.8}                       & \textbf{8.5}                          & \textbf{7.4}                         \\ \bottomrule
\end{tabular}

}

\caption{The faithfulness evaluation results for self-reflective rationales.}
\label{tab:faith}
\end{table*}

\looseness=-1
\paragraph{Baselines} 
We compare with the following approaches:
(1) Direct prompting for confidence extraction (\textbf{DP}): 
We directly ask the vanilla LLMs to give a confidence score from 1 to 10 in their previous response~\cite{tian2023just}.
(2) Self-consistency-based confidence estimate (\textbf{SC}): We use the self-consistency-based approach to derive the confidence estimates of LLMs.
Confidence is calculated as the ratio of response frequency to the number of samples~\cite{xiong2023can}.
(3) Prompting for correctness (\textbf{PC}): We ask the vanilla LLMs to judge whether their responses are correct or not~\citep{kadavath2022language}.
(4) \textbf{R-Tuning}: We train LLMs to generate binary confidence estimates (sure \textit{vs.} unsure) using a model-specific dataset~\cite{zhang2023r}.
(5) Aligning with self-consistency-based confidence (\textbf{AS}): We train LLMs to generate the confidence estimates derived from self-consistency prompting~\citep{yang2023alignment}.
(6) Grouping-based confidence estimates for calibration training (\textbf{GCE}): We group the samples in HotpotQA via clustering, and use the accuracy of samples in the group as the confidence estimates for all samples within that group. The constructed dataset is thus used for fine-tuning~\cite{lin2022teaching}.
We implement the baseline approaches and \approach on Mistral-7B~\cite{jiang2023mistral} for fair comparison.
To prove that \approach can generalize on multiple models, we also implement the baseline approaches and \approach on Llama 3 8B~\cite{dubey2024llama3herdmodels} in Appendix \ref{sec:llama}.



%


\subsection{Main Experimental Results}
\looseness=-1
\textbf{Confidence Calibration Performance.}
We show the ECE results (Table~\ref{tab:ece}) and the AUROC results (Table~\ref{tab:auroc} in the Appendix) that measure the correlation between the expressed confidence and the actual performance.
We observe that \approach significantly outperforms all baseline approaches in reducing the calibration error (ECE) and improving the distinction of confidence in correct and incorrect responses (AUROC). 
This conclusion holds in both in-distribution (HotpotQA) and out-of-distribution datasets, which demonstrates the general applicability of \approach.
%
Also, the difference of \approach from other baselines is mostly statistically significant ($p<0.05$), further demonstrating its capability to provide effective confidence estimates.

\noindent \textbf{Task Performance.}
We show the accuracy results in Table~\ref{tab:acc}.
\textbf{SC}, which uses multiple sampling, achieves overall better performance compared to other approaches. 
However, this results in high inference latency.
Compared to other baseline approaches, \approach can overall maintain the original task performance. 
This indicates that the task of confidence estimates doesn't conflict with the original task, consistent with previous work~\cite{Chen2022ACL, zhang2023r}.

%
%

\noindent \textbf{Faithfulness of the Generated Self-Reflective Rationales.}
The evaluation prompt for GPT-4 is shown in Appendix~\ref{app:prompt}.
We show the faithfulness results in Table~\ref{tab:faith}.
Due to the budget limits for GPT-4 evaluation,
we sample 100 instances from each dataset for evaluation. 
The instances with multiple selective reasoning chains are chosen for priority.
For all baseline approaches, we explicitly instruct LLMs to provide reasoning for the confidence levels assigned to their prior responses.
The results for DP and SC are combined, as both methods sample from the same LLM.

The experimental results show that \approach can generate more reasonable self-reflective rationales that indicate the internal uncertainty in LLMs as evidenced by inconsistencies across multiple sampled responses.
One exception is the StrategyQA dataset, which only contains True/False questions.
Consequently, typically only one or two responses are selected for each question, resulting in high variance in the evaluation.

\looseness=-1
We conduct human annotations to justify the use of GPT-4 for automatic evaluation. The details are described in Appendix~\ref{sec:human}.
We observe a Spearman's rank correlation coefficient of 0.89 between the ratings given by GPT-4 and humans, which demonstrates the reliability of automatic evaluation.

%

%
%

\subsection{Ablation Study}
We conduct an ablation study to verify several design choices in \approach:
(1) \textbf{w/o RL}: We evaluate \approach without the reinforcement learning from the task supervision stage.
(2) \textbf{w/o R \& CE}: We evaluate \approach that directly trains LLMs on the golden answer without the self-reflective rationales and confidence estimates in the supervised fine-tuning stage.
(3) \textbf{w/o R}: We evaluate \approach that directly trains LLMs on the golden answer and confidence estimates without the self-reflective rationales in the supervised fine-tuning stage.
(4) \textbf{w/ Naive RF}: We verify the effectiveness of the defined reward function in \approach. 
We compare with a simple intuitive reward function:
$
R = \mathbb{I}(\text{correct}) \times \text{confidence level} - \mathbb{I}(\text{incorrect}) \times \text{confidence level}
$.

The results are shown in Table~\ref{tab:ece}, Table~\ref{tab:base_ece}, Table~\ref{tab:auroc} (Appendix), and Table~\ref{tab:acc} for direct comparison with \approach.
Compared with \approach w/o RL, 
our results indicate that while supervised fine-tuning can enable LLMs to express calibrated confidence to a certain extent, incorporating RL with task-specific supervision further enhances the accuracy of these confidence estimates.
The ablation of the reward function also justifies our design choice in the RL stage.
For the supervised fine-tuning stage, both the self-reflective rationales and the confidence estimates contribute significantly to the calibrated confidence estimates.
Overall, the ablation results verify the effectiveness of all components in \approach.

 \begin{figure*}[t!]
\centering
\includegraphics[width=\textwidth]{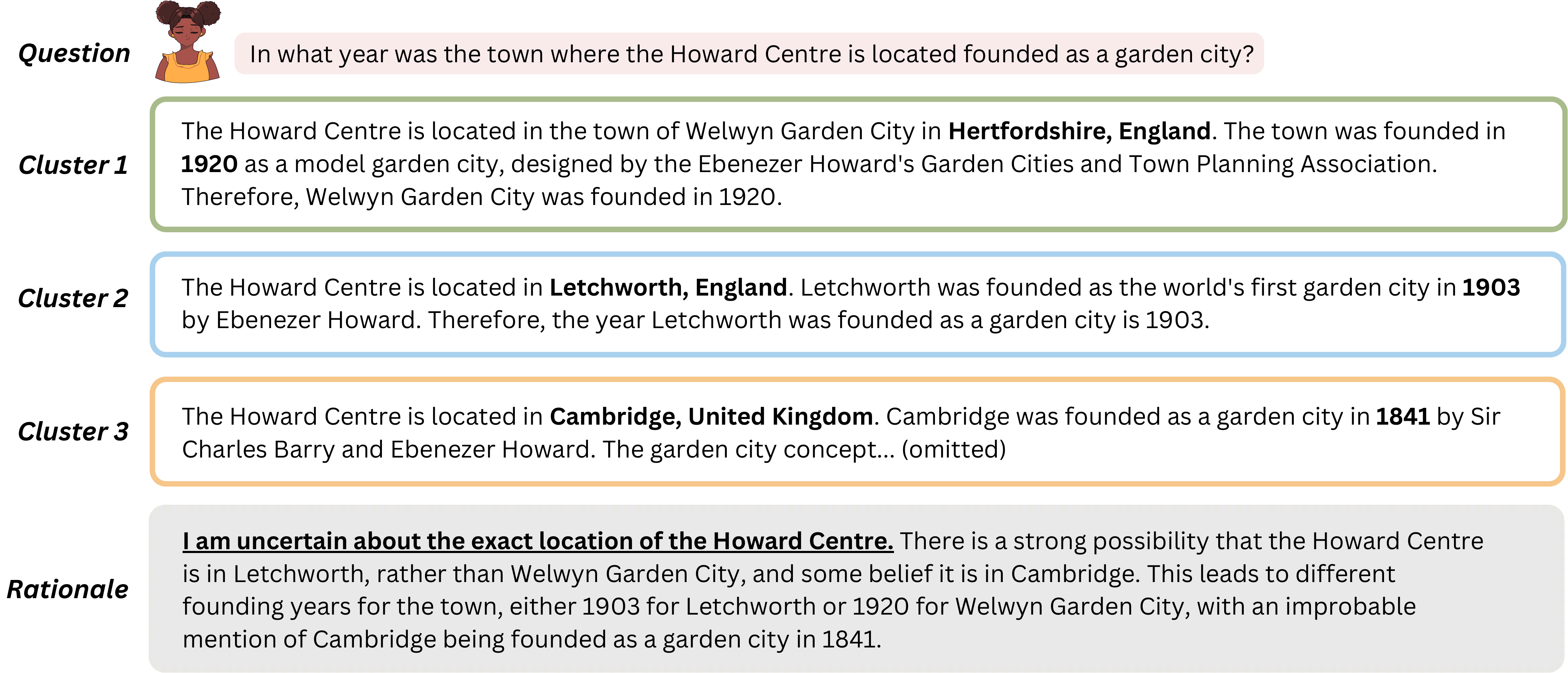}
 \caption{Case study of \approach's capability to generate insightful self-reflective rationales that effectively capture the internal uncertainty in LLMs. Various clusters illustrate a selection from 100 sampled responses, and the rationale is generated by LLMs. Another example is given in Figure~\ref{fig:casestudy2} in the Appendix.
 }
 \vspace{-5pt}
 \label{fig:casestudy}
 \end{figure*}

\begin{table}[t!]
\centering
\renewcommand\arraystretch{0.8}
\setlength{\tabcolsep}{9pt} 
\resizebox{0.45\textwidth}{!}{
\begin{tabular}{l|ccc}
\toprule
         & \multicolumn{1}{l}{\textbf{Unanswerable}} & \multicolumn{1}{l}{\textbf{Answerable}} & \multicolumn{1}{l}{ \textbf{$\ \ \ \Delta$}} \\ \midrule
DP       & 0.6696                                          & 0.7646                                        & 0.0950                              \\
SC       & 0.2317                                          & 0.3561                                        & 0.1244                             \\
PC       & 0.8676                                          & 0.9974                                        & 0.1298                             \\
R-Tuning & 0.7614                                          & 0.8210                                         & 0.0596                             \\
AS       & 0.4381                                          & 0.4503                                        & 0.0122                            \\
GCE      & 0.2987                                          & 0.2991                                        & 0.0004                             \\
\approach  & 0.4962                                          & 0.7406                                        & \textbf{0.2444}                             \\ \bottomrule
\end{tabular}

}
\caption{The confidence in unanswerable and answerable subsets of SQUADRUN.}
\vspace{-16pt}
\label{tab:unanswerable}
\end{table}
\subsection{Unanswerable Questions}
We measure whether LLMs demonstrate low confidence in responding to unanswerable questions, which serves as a clear indicator of their ability to accurately delineate their knowledge boundaries 
We choose the SQUADRUN dataset~\citep{rajpurkar2018know}, which contains both answerable and unanswerable questions. 
We measure the average confidence in the answerable and unanswerable subsets (see Table~\ref{tab:unanswerable}). 
We show that \approach enables LLMs to significantly reduce the confidence in unanswerable questions while maintain the confidence in the answerable parts, achieving the best confidence gap ($\Delta$) between the two subsets.

\subsection{Case Study}
\looseness=-1
We perform a case study to better understand our approach (see Figure~\ref{fig:casestudy} and Figure~\ref{fig:casestudy2} in the Appendix). 
We choose two straightforward questions from HotpotQA and prompt LLMs trained via \approach to generate the self-reflective rationales. 
Then we perform multiple sampling (100 times) and clustering to get a selection of representative responses.
These examples demonstrate \approach’s strong ability to detect and summarize internal uncertainties.
For example, in the first case, \approach expresses uncertainty about the exact location of the Howard Centre, identifying strong indications that it is likely in Letchworth and not Welwyn Garden City, with Cambridge being an unlikely option. 
This rationale acknowledges the mixed information leading to different founding years based on the location—1903 for Letchworth and 1920 for Welwyn Garden City, dismissing the 1841 Cambridge claim as highly improbable.
This capability for self-reflective generation has a profound impact on improving the reliability of LLM-based systems.
%



\section{Conclusion}
This paper presents a training framework \approach for eliciting more accurate and fine-grained confidence estimates and self-reflective rationales from LLMs.
\approach involves supervised finetuning with a model-specific dataset constructed by summarizing the difference between multiple reasoning chains and reinforcement learning with a properly designed reward function. 
Our evaluations across diverse datasets confirm that \approach reduces calibration errors, maintains performance, and generates insightful rationales. 



\section*{Limitations}
A potential limitation of \approach is its dependence on multiple sampled chains of reasoning to develop self-reflective rationales for training.
There is still an ongoing debate regarding the faithfulness of CoT reasoning, specifically questioning whether it authentically represents the thinking process of LLMs~\citep{lanham2023measuring, bentham2024chain, turpin2024language}. 
The unfaithful CoT reasoning can cause unfaithful self-reflective rationales.
Nonetheless, our ablation study demonstrates that these self-reflective rationales substantially enhance calibration performance. Further improvements in the effectiveness and faithfulness of \approach could potentially be achieved by integrating methods from recent research aimed at increasing the faithfulness of CoT reasoning, as suggested by \citet{lyu2023faithful}.

%

\section*{Ethical Considerations}
This work aims to improve the performance of LLMs in eliciting more fine-grained confidence estimates and self-reflective rationales. In the case of this work, it involves the use of Mistral 7B and GPT-4, so the same risks from LLMs research are also applicable to this work \cite{10.1145/3442188.3445922}.

While \approach aims to enhance trust in AI by providing clear confidence expressions and self-reflective rationales, there is a risk that users might over-rely on these confidence estimates. If the self-reflective rationales are not accurate or fail to capture the true uncertainty of the model, it could lead to potentially harmful decisions based on the model’s outputs. Therefore, users are advised to check important information before making crucial decisions.

This paper works on several publicly available datasets including HotpotQA, TruthfulQA, StrategyQA, FEVER, HaluEval, and ParaRel. They are available for the research community to study under Apache 2.0, Apache 2.0, MIT, CC-BY-SA 3.0, Apache 2.0, and MIT licenses respectively. Data is anonymized, thus our work does not propagate any privacy problems about any specific entities.

Finally, we carried out human annotations for analysis purposes. Since the amount of work is small, we and the annotators agree to consider it as a voluntary service. We have sufficiently discussed the specific use of the annotations and potential risks to annotators before the work.

\section*{Acknowledgments}
This work is supported in part by the US National Science Foundation under grant NSF-IIS2226108. Any opinions, findings, and conclusions or recommendations expressed in this material are those of the author(s) and do not necessarily reflect the views of the National Science Foundation.

\bibliography{main}
\ 

\ 

\

\newpage
\appendix
\section*{Appendix}
\section{Prompt}
\label{app:prompt}

\subsection{GPT-4 Summarization Prompt}
\begin{lstlisting}
Your task is to analyze a question provided to you along with a set of correct and incorrect responses generated by my model. Your objective is to identify and summarize the inconsistency in the models' responses that can explain why my model is uncertain about the correct answer.

Please note that:
1. You should give the reasons from a first-person perspective, as if you are my model that gives the provided responses and confidence scores.
2. Limit your explanation to the knowledge and facts the model possesses about the question.
3. Keep your summary brief, aiming for 1-3 sentences.
4. Each response is paired with a confidence score at the beginning. Include the confidence score that accompanies each response in your summary.
5. Please directly provide the summarized reason without any greetings or other unnecessary information. If you find the incorrect responses are consistent with the correct response regarding the question, please directly return N/A. 
6. Importantly, my model doesn't have access to the ground truth. Therefore, the summarized reason should not have any statement about correctness or incorrectness of the responses. You should only focus on discussing the uncertainty in the knowledge and facts based on the inconsistency in the responses.
7. Importantly!! My model only has access to the correct response. Thus, the summary should not include any statement like "My different responses have...", "my responses about ...", "the multiple responses...", etc. You should not say "my responses" or "the responses" anywhere in the summary.

Here is an example:
Question: Sky High starred the actress who is married to which actor?

Correct Response:
(6% confidence) The actress who starred in "Sky High" (2005) and is married to an actor is Kelly Preston. Her husband is John Travolta. The two have been married since 1991 and have three children together.

Incorrect Responses:
1. (33% confidence) The actress who starred in "Sky High" (2005) and is married to an actor is Kristen Bell. Bell voiced the main character, Layla, in "Sky High," and she is married to Dax Shepard, who is also an actor.
2. (17% confidence) The actress who starred in "Sky High" (2005) and is married to an actor is Kelly Clarkson. Her acting debut was in this film, and she married singer and actor Brandon Blackstock in 2013.

The output can be: I am uncertain about the correct actress in "Sky High". There is a high probability that the actress is Kristen Bell, instead of Kelly Preston.  I am confused about her voice acting roles with on-screen appearances. There is also some probability that the actress is Kelly Clarkson.

Now consier the following case:
Question: {}

Correct Response: 
{}

Incorrect Responses:
{}

\end{lstlisting}

\begin{table*}[t!]
\centering
\renewcommand\arraystretch{0.8}
\setlength{\tabcolsep}{9pt} 

\resizebox{0.99\textwidth}{!}{
\begin{tabular}{l|cccccc}
\toprule
Method | Dataset  & HotpotQA      & TruthfulQA      & StrategyQA                    & FEVER           & HaluEval        & ParaRel         \\ \midrule
DP                & 0.3222          & 0.5667          & 0.5193                        & 0.5371          & 0.5278          & 0.5291          \\
SC                & 0.5765          & 0.4939          & {\color[HTML]{000000} 0.5498} & 0.5472          & 0.5843          & 0.5546          \\
PC                & 0.6636	 & 0.5244 &	0.5385 &	0.5139	 & 0.5544 & 	0.6181        \\
R-Tuning          &  0.6529	      &  0.5980              & 	0.5406                                   & 	0.5688               &         	0.5330	       &    0.5424                    \\
AS       & 0.4955      & 	0.4835	  	& 0.5391   &	0.5101	    & 0.5569	                        & 	0.5713         \\  
GCE               &  0.5042	               &          0.4966	     &  0.5043	                              &    0.4942           &     	0.4907        &     	0.5031                   \\ \midrule
\approach           & \textbf{0.7156} & \textbf{0.6107} & \textbf{0.6074}               & \textbf{0.6481} & \textbf{0.7318} & \textbf{0.6816} \\
w/o RL            & 0.6524          & 0.5675          & 0.5910                        & 0.5798          & 0.5929          & 0.6003          \\
w/o R \& CE  & 0.5256          & 0.5724          & 0.5738                        & 0.6059          & 0.6002          & 0.5823          \\ 
w/o R   & 0.4928 &	0.4952	 & 0.4567 &	0.4831 &	0.4893 &	0.4853        \\  
w/ Naive RF & 0.5140 &	0.4907 &	0.5091 & 	0.5137	 & 0.5147 &	0.5053        \\ 
\bottomrule
\end{tabular}

}
\caption{The AUROC evaluation results of baselines, \approach, and various ablations.}

\label{tab:auroc}
\end{table*}
\begin{table*}[t!]
\centering
\renewcommand\arraystretch{0.8}
\setlength{\tabcolsep}{9pt} 
\resizebox{0.80\textwidth}{!}{
\begin{tabular}{l|cc|cc|cc}
\toprule
\multirow{2}{*}{Method | Dataset} & \multicolumn{2}{c|}{HotpotQA} & \multicolumn{2}{c|}{FEVER} & \multicolumn{2}{c}{HaluEval}  \\
& ECE & Accuracy  & ECE & Accuracy  & ECE & Accuracy    \\
\midrule
DP & 0.4744 & \textbf{0.4762} & 0.6167 & 0.6282 & 0.7750 & 0.8884\\
SC & 0.4798 & 0.3475 & 0.4775 & \textbf{0.6325} & 0.5220 & 0.7168 \\
PC & 0.4604 & 0.3678 & 0.4513 & 0.6062 & 0.3375 & 0.8159 \\
R-Tuning & 0.4094 & 0.4348 & 0.4250 & 0.5588 & 0.3393 & 0.6604 \\
AS & 0.3525 & 0.4230 & 0.3944 & 0.4937 & 0.3027 & \textbf{0.9150} \\
GCE & 0.4056 & 0.3480 & 0.4330 & 0.5860 & 0.3190 & 0.8650 \\
\approach & \textbf{0.3296} & 0.4284 & \textbf{0.3844} & 0.6125 & \textbf{0.2427} & 0.8552\\
\bottomrule
\end{tabular}
}
\caption{The ECE and accuracy evaluation results of baselines and \approach on a different base model, Llama 3 8B.}
\vspace{-5pt}

\label{tab:base_ece}
\end{table*}

 \begin{figure*}[t!]
\centering
\includegraphics[width=\textwidth]{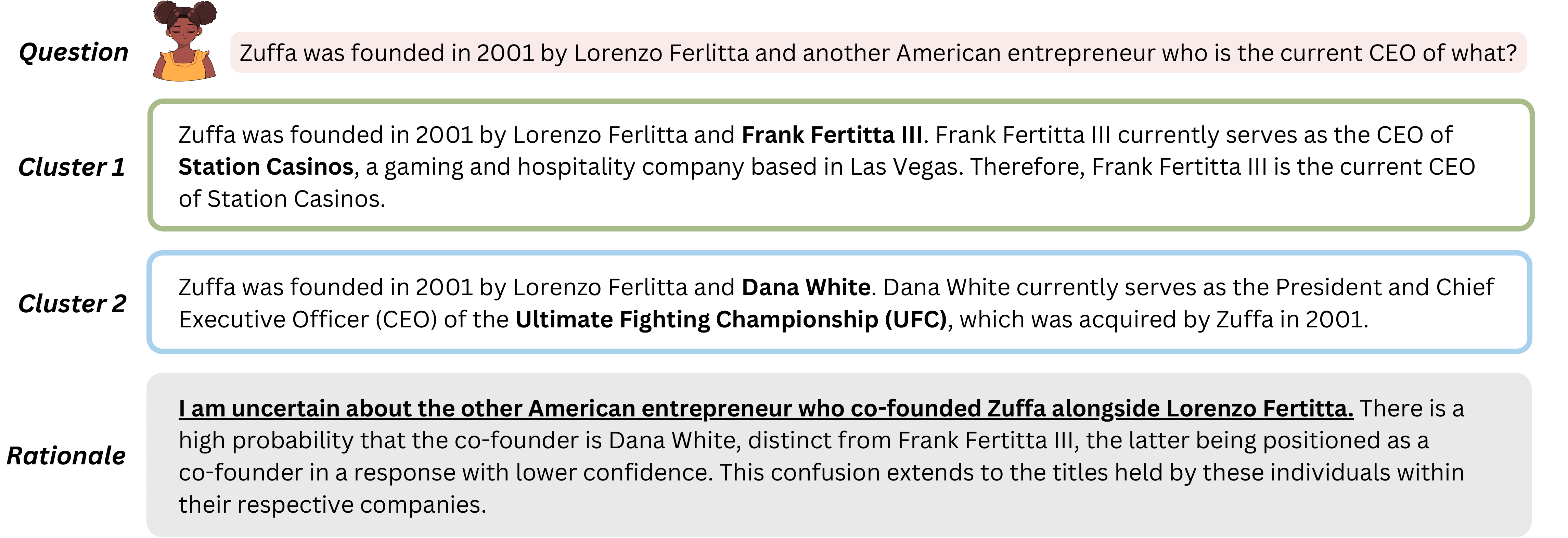}
 \caption{Case study of \approach's capability to generate insightful self-reflective rationales.
 }
 \label{fig:casestudy2}
 \end{figure*}

\subsection{GPT-4 Evaluation Prompt}
\begin{lstlisting}
Your task is to analyze whether a summarized explanation correctly explains the inconsistency in multiple sampled responses to a question. 
Note that each response is paired with a proportion at the beginning, indicating the frequency of the response in the sampled responses. You should output a score from 0 to 10, indicating the quality of the explanation.
You should first provide your reasoning for the correctness of the explanation, and then assign a score based on the quality of the explanation. The output should be in the following format: reason: [REASON] score: [SCORE].

Here is an example:
Question: Sky High starred the actress who is married to which actor?

Sampled Responses:
1. (6%) The actress who starred in "Sky High" (2005) and is married to an actor is Kelly Preston. Her husband is John Travolta. The two have been married since 1991 and have three children together.
2. (33%) The actress who starred in "Sky High" (2005) and is married to an actor is Kristen Bell. Bell voiced the main character, Layla, in "Sky High," and she is married to Dax Shepard, who is also an actor.
3. (17%) The actress who starred in "Sky High" (2005) and is married to an actor is Kelly Clarkson. Her acting debut was in this film, and she married singer and actor Brandon Blackstock in 2013.

Reason: I am uncertain about the correct actress in "Sky High". There is a high probability that the actress is Kristen Bell, instead of Kelly Preston.  I am confused about her voice acting roles with on-screen appearances. There is also some probability that the actress is Kelly Clarkson.

Then your output can be: 
reason: The provided reason is clear and well-structured. The explanation is logical and provides a clear rationale for the uncertainty expressed in the sampled responses. score: 9


Now consier the following case:
Question: {}

Sampled Responses:
{}

Reason: {}


\end{lstlisting}

\section{Empirical Evidence}
\label{sec:appb}
\looseness=-1
In our study, we conduct both quantitative and qualitative analyses of the diversity of reasoning chains within each clustering group.
The quantitative analysis reveals an average similarity of 0.94 across the reasoning chains, indicating high consistency and similarity within each cluster. 
For qualitative analysis, we randomly pick 3 clusters and manually inspect the reasoning chains in each cluster. 
We discover that the exact similarity rates within these clusters are 58\%, 80\%, and 74\%, respectively, with the variations primarily involving minor differences in wording and sentence structure in the remaining reasoning chains.
This supports our design decision to select one instance per cluster at random.
%


\section{Human Annotations for GPT-4 Evaluation}
\label{sec:human}
We randomly select 200 questions from multiple test datasets, providing each question along with corresponding reasoning chains and self-reflective rationales to two human annotators. The two human annotators are PhD students. These annotators, guided by the instructions for GPT-4, are asked to rate each case on a scale from 1 to 10, and the final score for each case is the average over two human annotators. 
We measure the pearman’s rank correlation coefficient between human evaluation and GPT-4 evaluation. 
The correlation coefficient is 0.89, which proves the reliability of automatic evaluation using GPT-4. 
\section{Experiments of \approach on Llama}
\label{sec:llama}

To test the generalizability of \approach, we test \approach on a different base model, Llama 3 8B. The baselines are the same as mentioned in Section \ref{sec:evaluation_setting}. The experiment results are given in Table \ref{tab:base_ece}. Our results indicate that \approach can generalize in different base models and has superior performance over other baseline methods in ECE, without a significant loss in accuracy.

\end{document}